\documentclass{article}

\usepackage{ijcai13}

\usepackage{times}
\usepackage{helvet}
\usepackage{courier}

\usepackage{amsmath}
\usepackage{amssymb}

\usepackage{latexsym}
\usepackage{picinpar}

\newtheorem {definition}{Definition}

\newtheorem {theorem}{Theorem}

\newtheorem {example}{Example}

\begin{document}

\title{Logical Fuzzy Preferences}
\author{ Emad Saad \\
emsaad@gmail.com
}

\maketitle
\begin{abstract}

We present a unified logical framework for representing and reasoning about both quantitative and qualitative preferences in fuzzy answer set programming \cite{Saad_DFLP,Saad_eflp,Subrahmanian_B}, called fuzzy answer set optimization programs. The proposed framework is vital to allow defining quantitative preferences over the possible outcomes of qualitative preferences. We show the application of fuzzy answer set optimization programs to the course scheduling with fuzzy preferences problem described in \cite{Saad_DFLP}. To the best of our knowledge, this development is the first to consider a logical framework for reasoning about quantitative preferences, in general, and reasoning about both quantitative and qualitative preferences in particular.

\end{abstract}

\section {Introduction}

Fuzzy reasoning is vital in most real-world applications. Therefore, developing well-defined frameworks for representing and reasoning in the presence of fuzzy environments is inevitable. Thus many frameworks have been proposed for fuzzy reasoning. Among these frameworks are fuzzy answer set programming \cite{Saad_DFLP,Saad_eflp,Subrahmanian_B}, which are fuzzy logic programs with fuzzy answer set semantics \cite{Saad_DFLP,Saad_eflp,Subrahmanian_B}.

As noted in \cite{Saad_DFLP}, the importance of the fuzzy answer set programming frameworks of \cite{Saad_DFLP,Saad_eflp,Subrahmanian_B} lies in the fact that the fuzzy answer set programming frameworks of \cite{Saad_DFLP,Saad_eflp,Subrahmanian_B} are strictly more expressive than the fuzzy answer set programming framework of \cite{Janssen_1,Janssen_2}. This is because the way how a rule is fired in \cite{Saad_DFLP,Saad_eflp,Subrahmanian_B} is close to the way how it fires in classical answer set programming \cite{Gelfond_A,Gelfond_B}, which makes any possible extension to \cite{Saad_DFLP,Saad_eflp,Subrahmanian_B} to more expressive forms of fuzzy answer set programming is more intuitive and more flexible.

In \cite{Saad_DFLP}, an expressive fuzzy answer set programming framework has been developed, namely extended and normal disjunctive fuzzy logic programs with fuzzy answer set semantics, that generalize and subsume; classical extended and classical normal disjunctive logic programs with classical answer set semantics \cite{Gelfond_B,Dix}; extended fuzzy logic programs with fuzzy answer set semantics \cite{Saad_eflp}; and normal fuzzy logic programs with fuzzy answer set semantics \cite{Subrahmanian_B}, in a unified logic programming framework to allow non-monotonic negation, classical negation, and disjunctions under fuzzy uncertainty.

The fuzzy answer set programming framework of \cite{Saad_DFLP} is necessary to provide the ability to assign fuzzy uncertainly over the possible outcomes of qualitative uncertainty, which is required in many real-world applications, e.g., representing and reasoning about preferences in fuzzy environments. In a unified logical framework, \cite{Saad_DFLP} allows directly and intuitively to represent and reason in the presence of both fuzzy uncertainty and qualitative uncertainty. This has been illustrated by applying the fuzzy answer set programming framework of \cite{Saad_DFLP} to the course scheduling with fuzzy preferences problem \cite{Saad_DFLP}, where an instructor preferences over courses are represented as a fuzzy set over courses, instructor preferences over class rooms are represented as a fuzzy set over class rooms, and instructor preferences over time slots are represented as a fuzzy set over time slots. The fuzzy answer set program encoding of the course scheduling with fuzzy preferences problem in \cite{Saad_DFLP} provided all possible solutions to the problem represented by fuzzy answer sets of the fuzzy answer set program encoding of the problem.

For example, consider this simple instance of the course scheduling with fuzzy preferences problem described in \cite{Saad_DFLP}. Assume that one of two courses, $c_1, c_2$, need to be assigned to an instructor $i$ such that instructor $i$ is assigned exactly one course. Consider instructor $i$ prefers to teach $c_1$ over $c_2$, where this preference relation is specified as a fuzzy set over the courses $c_1, c_2$. Consider instructor $i$'s preference in teaching $c_1$ is characterized by the grade membership value $0.3$ and instructor $i$'s preference in teaching $c_2$ is characterized by the grade membership value $0.5$. Thus, this course scheduling with fuzzy preferences problem instance can be encoded as a fuzzy answer set program (a disjunctive fuzzy logic program with fuzzy answer set semantics) of the form \[teaches(i, c_1):0.3  \; \vee \; teaches(i, c_2):0.5 \] with $\{teaches(i, c_1):0.3 \}$ and $\{teaches(i, c_2):0.5 \}$ are the fuzzy answer sets of the program, according to the fuzzy answer set semantics of fuzzy answer set programming of \cite{Saad_DFLP}.

It is clear that the fuzzy answer set $\{teaches(i, c_2):0.5 \}$ encodes instructor $i$'s top teaching preferences, which implies that the fuzzy answer set $\{teaches(i, c_2):0.5 \}$ is the most preferred fuzzy answer set according to the preferences ({\em quantitative preferences}) encoded by the fuzzy answer set program. In addition, consider instructor $i$ is neutral regarding teaching either course, where $i$'s this teaching preference is characterized by the grade membership value $0.3$ for both courses. In this case, this course scheduling with fuzzy preferences problem instance can be encoded as a fuzzy answer set program of the form \[teaches(i, c_1):0.3 \; \vee \; teaches(i, c_2):0.3\] with $\{teaches(i, c_1):0.3 \}$ and $\{teaches(i, c_2):0.3 \}$ are the fuzzy answer sets, according to the fuzzy answer set semantics of fuzzy answer set programming of \cite{Saad_DFLP}. Although instructor $i$ is neutral regarding teaching either course with preference $0.3$ each, however, it can be the case that instructor $i$ has more appeal in teaching course $c_1$ over course $c_2$ ({\em qualitative preferences}). This makes $\{teaches(i, c_1):0.3 \}$ is the most preferred fuzzy answer set in this case.

The current semantics of fuzzy answer set programs \cite{Saad_DFLP,Saad_eflp,Subrahmanian_B} does not have the ability to rank fuzzy answer sets neither according to quantitative preferences nor according to qualitative preferences. Rather, fuzzy answer set programs semantics is capable of determining fuzzy answer sets that {\em satisfy} quantitative preferences presented by the fuzzy answer set program and considers all the resulting fuzzy answer sets as equally preferred.

However, for many applications, it is necessary to rank the fuzzy answer sets generated by the fuzzy answer set programs from the top (most) preferred fuzzy answer set to the least preferred fuzzy answer set, where the top (most) preferred fuzzy answer set is the one that is most desirable. This requires fuzzy answer set programs to be capable of representing both quantitative and qualitative preferences and to be capable of reasoning in the presence of both quantitative and qualitative preferences across fuzzy answer sets.

In this paper we develop a unified logical framework that is capable of representing and reasoning about both quantitative and qualitative preferences. This is achieved by defining the notion of {\em fuzzy answer set optimization programs}. Fuzzy answer set optimization programs modify and generalize the classical answer set optimization programs described in \cite{ASO}. We show the application of fuzzy answer set optimization programs to the course scheduling with fuzzy preferences problem, where a fuzzy answer set program \cite{Saad_DFLP} (disjunctive fuzzy logic program with fuzzy answer set semantics) is used as fuzzy answer sets generator rules. To the best of our knowledge, this development is the first to consider a logical framework for reasoning about quantitative preferences, in general, and reasoning about both quantitative and qualitative preferences in particular.

Fuzzy answer set optimization programs are fuzzy logic programs under the fuzzy answer set semantics whose fuzzy answer sets are ranked according to fuzzy preference relations specified by the user. A fuzzy answer set optimization program is a union of two sets of fuzzy logic rules, $\Pi =  R_{gen} \cup R_{pref}$. The first set of fuzzy logic rules, $R_{gen}$, is called the generator rules that generate the fuzzy answer sets that satisfy every fuzzy logic rule in $R_{gen}$. $R_{gen}$ is any set of fuzzy logic rules with well-defined fuzzy answer set semantics including normal, extended, and disjunctive fuzzy logic rules \cite{Saad_DFLP,Saad_eflp,Subrahmanian_B}, as well as fuzzy logic rules with fuzzy aggregates (all are forms of {\em fuzzy answer set programming}). The second set of fuzzy logic rules, $R_{pref}$, is called the {\em fuzzy preference rules}, which are fuzzy logic rules that represent the user's {\em quantitative} and {\em qualitative} preferences over the fuzzy answer sets generated by $R_{gen}$. The fuzzy preferences rules in $R_{pref}$ are used to rank the generated fuzzy answer sets from $R_{gen}$ from the top preferred fuzzy answer set to the least preferred fuzzy answer set. Similar to \cite{ASO}, an advantage of fuzzy answer set optimization programs is that $R_{gen}$ and $R_{pref}$ are independent. This makes fuzzy preference elicitation easier and the whole approach is more intuitive and easy to use in practice.

\section{Fuzzy Answer Set Semantics}

Since we use fuzzy logic rules under the fuzzy answer set semantics to generate fuzzy answer sets, that are ultimately ranked by fuzzy preference rules, in this section we recall the fuzzy answer set semantics of disjunctive fuzzy logic sets of rules, a form of fuzzy answer set programming, as presented in \cite{Saad_DFLP}.

\subsection{Syntax}

Consider a first-order language $\cal L$ with finitely many predicate symbols, function symbols, constants, and infinitely many variables. The Herbrand base of $\cal L$ is denoted by $\cal {B_L}$. Negation as failure or non-monotonic negation is denoted by $not$. The grade membership values are assigned to atoms in $\cal {B_L}$ as values from $[0,1]$. A \emph{fuzzy annotation}, $\mu$, is either a constant (\emph{fuzzy annotation constant}) in $[0, 1]$, a variable (\emph{fuzzy annotation variable}) ranging over $[0, 1]$, or $f(\mu_1,\ldots,\,\mu_n)$ (\emph{fuzzy annotation function}) where $f$ is a representation of a computable function $f: ([0, 1])^n \rightarrow [0, 1]$ and $\mu_1,\ldots, \mu_n$ are fuzzy annotations. A disjunctive fuzzy logic rule is an expression of the form

\begin{eqnarray}
A_1:\mu_1 \; \vee \ldots \vee \; A_k:\mu_k \leftarrow  A_{k+1}:\mu_{k+1},  \ldots, \notag \\
A_m:\mu_m, not\; A_{m+1}:\mu_{m+1}, \notag \\
\ldots, not\;A_{n}:\mu_{n}, \label{rule}
\end{eqnarray}
where $\forall (1 \leq i \leq n)$, $A_i$ is an atom and $\mu_i$ is a fuzzy annotation.

Intuitively, a disjunctive fuzzy logic rule means that if it is \emph{believable} that the grade membership value of each $(k+1 \leq i \leq m)$ $A_i$ is at least $\mu_i$ and it is \emph{not believable} that the grade membership value of each $(m+1 \leq j \leq n)$ $A_j$ is at least $\mu_j$, then there exist at least $(1 \leq i \leq k)$ $A_i$ such that the grade membership value of $A_i$ is at least $\mu_i$.

A disjunctive fuzzy logic rule is ground if it does not contain any variables.

\subsection{Fuzzy Answer Sets Semantics}

A fuzzy interpretation, $I$, of a set of disjunctive fuzzy logic rules is a fuzzy set in the Herbrand base, $\cal {B_L}$, whose grade membership function is a mapping ${\cal B_L} \rightarrow [0, 1]$. This implies that a fuzzy interpretation, $I$, is the mapping $I: {\cal B_L} \rightarrow [0, 1]$, where the grade membership value of an atom, $A \in {\cal B_L}$, in $I$, is $I(A)$. Let $r$ be a disjunctive fuzzy logic rule of the form (\ref{rule}). Let $head(r) = A_1:\mu_1 \; \vee \ldots \vee \; A_k:\mu_k$ and \\ $body(r) = A_{k+1}:\mu_{k+1},  \ldots, A_m:\mu_m, not\; A_{m+1}:\mu_{m+1},\ldots, not\;A_{n}:\mu_{n}$.

\begin{definition}
\label{def:prop_sat} Let $R$ be a set of ground disjunctive fuzzy logic rules, $I$ be a fuzzy interpretation of $R$, and $r$ be a disjunctive fuzzy logic rule of the form (\ref{rule}). Then:

\begin{list}{$\bullet$}{\topsep=1pt \parsep=0pt \itemsep=1pt}

\item $I$ satisfies $A_i : \mu_i$ in $head(r)$ iff $\mu_i \leq I(A_i)$.

\item $I$ satisfies $A_i : \mu_i$ in $body(r)$ iff $\mu_i \leq I(A_i)$.

\item $I$ satisfies $not\; A_j : \mu_j$ in $body(r)$ iff $\mu_j \nleq I(A_j)$.

\item $I$ satisfies $body(r)$ iff $\forall(k+1 \leq i \leq m)$, $I$ satisfies $A_i : \mu_i$ and $\forall(m+1 \leq j \leq n)$, $I$ satisfies $not\; A_j : \mu_j$.

\item $I$ satisfies $head(r)$ iff $\exists  i$ $(1 \leq i \leq k)$ such that $I$ satisfies $A_i : \mu_i$.

\item $I$ satisfies $r$ iff $I$ satisfies $head(r)$ whenever $I$ satisfies $body(r)$ or $I$ does not satisfy $body(r)$.

\item $I$ satisfies $R$ iff $I$ satisfies every disjunctive fuzzy logic rule in $R$ and for every atom $A_i \in {\cal B_L}$, we have $\max \{\!\!\{\mu_i \; (1 \leq i \leq k)  \: | \: A_1:\mu_1  \vee \ldots \vee  A_k:\mu_k \leftarrow body(r) \in R$, $I$ satisfies $body(r)$, and $I$ satisfies $A_i:\mu_i \}\!\!\}\leq I(A_i).$

\end{list}
\label{def:fuzzy_sat}
\end{definition}
A fuzzy model of a set of disjunctive fuzzy logic rules, $R$, is a fuzzy interpretation for $R$ that satisfies $R$. A fuzzy model, $I$, of $R$ is called a minimal fuzzy model if there is no fuzzy model, $I'$, for $R$ such that $I' < I$. Let $R$ be a set of ground disjunctive fuzzy logic rules and $I$ be a fuzzy interpretation. Then, the fuzzy reduct, $R^I$, of $R$ w.r.t. $I$ is a set of non-monotonic-negation-free disjunctive fuzzy logic rules, $R^I$, where:
\begin{eqnarray*}
A_1 : \mu_1 \; \vee \ldots \; \vee \; A_k:\mu_k \leftarrow A_{k+1}:\mu_{k+1}, \ldots, \\ A_m : \mu_m
\in R^I
\end{eqnarray*}
iff
\begin{eqnarray}
A_1:\mu_1 \; \vee \ldots \vee \; A_k:\mu_k \leftarrow A_{k+1}:\mu_{k+1},  \ldots, \notag \\
A_m:\mu_m, not\; A_{m+1}:\mu_{m+1},\ldots, \notag \\
not\;A_{n}:\mu_{n} \in R, \notag
\end{eqnarray}
and $\forall (m+1 \leq j \leq n),\:  \mu_j \nleq I(A_j)$.

\begin{definition} A fuzzy interpretation, $I$, for a set of disjunctive fuzzy logic rules, $R$, is a fuzzy answer set of $R$ if $I$ is a minimal fuzzy model of $R^I$.
\end{definition}

\section{Fuzzy Answer Set Optimization Programs}

A fuzzy answer set optimization program is a union of two sets of fuzzy logic rules, $\Pi =  R_{gen} \cup R_{pref}$, where $R_{gen}$ is the set of the fuzzy answer sets generator rules and $R_{pref}$ is the set of the fuzzy preference rules. In our introduction of fuzzy answer set optimization programs, we focus on the syntax and semantics of the {\em fuzzy preference rules}, $R_{pref}$, of the fuzzy answer set optimization programs, since the syntax and semantics of the fuzzy answer sets generator rules, $R_{gen}$, are the same as syntax and semantics of any set of fuzzy logic rules with fuzzy answer set semantics as described in \cite{Saad_DFLP,Saad_eflp,Subrahmanian_B}.


\subsection{Fuzzy Preference Rules Syntax}

Let ${\cal L}$ be a first-order language with finitely many predicate symbols, function symbols, constants, and
infinitely many variables. A literal is either an atom $A$ or the negation of atom $A$ ($\neg A$), where $\neg$ is the classical negation. Non-monotonic negation or the negation as failure is denoted by $not$. ${\cal B_L}$ is the Herbrand base of ${\cal L}$. Let $Lit$ be the set of all literals in ${\cal L}$. A \emph{fuzzy annotation}, $\mu$, is either a constant (\emph{fuzzy annotation constant}) in $[0, 1]$, a variable (\emph{fuzzy annotation variable}) ranging over $[0, 1]$, or $f(\mu_1,\ldots,\,\mu_n)$ (\emph{fuzzy annotation function}), where $f$ is a representation of a computable function $f: ([0, 1])^n \rightarrow [0, 1]$ and $\mu_1,\ldots, \mu_n$ are fuzzy annotations.

If $l$ is a literal and $\mu$ is a fuzzy annotation, then $l:\mu$ is called a {\em fuzzy annotated literal}. Let $S$ be a set of fuzzy annotated literals. A boolean combination over $S$ is a boolean formula over fuzzy annotated literals in $S$ constructed by conjunction, disjunction, and non-monotonic negation ($not$), where non-monotonic negation is combined only with fuzzy annotated literals.

\begin{definition} A fuzzy preference rule, $r$, over a set of fuzzy annotated literals, $S$, is an expression of the form
\begin{eqnarray}
C_1 \succ C_2 \succ \ldots \succ C_k \leftarrow l_{k+1}:\mu_{k+1},\ldots, l_m:\mu_m, \notag \\
not\; l_{m+1}:\mu_{m+1},\ldots, not\;l_{n}:\mu_{n} \label{rule:pref}
\end{eqnarray}
where $l_{k+1}:\mu_{k+1}, \ldots, l_{n}:\mu_{n}$ are fuzzy annotated literals and $C_1, C_2, \ldots, C_k$ are boolean combinations over $S$.
\end{definition}
Let $head(r) = C_1 \succ C_2 \succ \ldots \succ C_k$ and $body(r) = l_{k+1}:\mu_{k+1},\ldots, l_m:\mu_m, not\; l_{m+1}:\mu_{m+1},\ldots, not\;l_{n}:\mu_{n}$, where $r$ is fuzzy preference rule of the form (\ref{rule:pref}). Intuitively, a fuzzy preference rule, $r$, of the form (\ref{rule:pref}) means that any fuzzy answer set that satisfies $body(r)$ and $C_1$ is preferred over the fuzzy answer sets that satisfy $body(r)$, some $C_i$ $(2 \leq i \leq k)$, but not $C_1$, and any fuzzy answer set that satisfies $body(r)$ and $C_2$ is preferred over fuzzy answer sets that satisfy $body(r)$, some $C_i$ $(3 \leq i \leq k)$, but neither $C_1$ nor $C_2$, etc.

\begin{definition} Formally, a fuzzy answer set optimization program is a union of two sets of fuzzy logic rules, $\Pi =  R_{gen} \cup R_{pref}$, where $R_{gen}$ is a set of fuzzy logic rules with fuzzy answer set semantics, the {\em generator} rules, and $R_{pref}$ is a set of fuzzy preference rules.
\end{definition}

\subsection{Fuzzy Preference Rules Semantics}

In this section, we define the satisfaction of fuzzy preference rules, and the ranking of the fuzzy answer sets with respect to a fuzzy preference rule and with respect to a set of fuzzy preference rules. We say that a set of fuzzy preference rules are ground if no variables appearing in any of its fuzzy preference rules.

\begin{definition} Let $\Pi = R_{gen} \cup R_{pref}$ be a ground fuzzy answer set optimization program, $I$ be a fuzzy answer set of $R_{gen}$ (possibly partial), and $r$ be a fuzzy preference rule in $R_{pref}$. Then the satisfaction of a boolean combination, $C$, appearing in $head(r)$, by $I$, denoted by $I \models C$, is defined inductively as follows:

\begin{itemize}

\item $I \models l:\mu$ iff  $\mu \leq I(l)$.

\item $I \models not\;l:\mu$ iff $\mu \nleq I(l)$ or $l$ is undefined in $I$.

\item $I\models C_1 \wedge C_2$ iff $I \models C_1$ and $I \models C_2$.

\item $I  \models C_1 \vee C_2$ iff $I \models C_1$ or $I \models C_2$.
\end{itemize}
Given $l_i:\mu_i$ and $not\;l_j:\mu_j$ appearing in  $body(r)$, the satisfaction of $body(r)$ by $I$, denoted by $I \models body(r)$, is defined inductively as follows:
\begin{itemize}
\item $I \models l_i:\mu_i$ iff $\mu_i \leq I(l_i)$

\item $I \models not\;l_j:\mu_j$ iff $\mu_j \nleq I(l_j)$ or $l_j$ is undefined in $I$.

\item $I \models body(r)$ iff $\forall(k+1 \leq i \leq m)$, $I \models l_i : \mu_i$ and $\forall(m+1 \leq j \leq n)$, $I \models not\; l_j : \mu_j$.
\end{itemize}

\end{definition}
The following definition specifies the satisfaction of the fuzzy preference rules.

\begin{definition} Let $\Pi = R_{gen} \cup R_{pref}$ be a ground fuzzy answer set optimization program, $I$ be a fuzzy answer set of $R_{gen}$, $r$ be a fuzzy preference rule in $R_{pref}$, and $C_i$ be a boolean combination in $head(r)$. Then, we define the following notions of satisfaction of $r$ by $I$:

\begin{itemize}
\item $I \models_{i} r$ iff $I \models body(r)$ and $I \models C_i$.

\item $I \models_{irr} r$ iff $I \models body(r)$ and $I$ does not satisfy any $C_i$ in $head(r)$.

\item $I \models_{irr} r$ iff $I$ does not satisfy $body(r)$.
\end{itemize}
\end{definition}
$I \models_{i} r$ means that $I$ satisfies the body of $r$ and the boolean combination $C_i$ that appears in the head of $r$. However, $I \models_{irr} r$ means that $I$ is irrelevant (denoted by $irr$) to $r$ or, in other words, $I$ does not satisfy the fuzzy preference rule $r$, because either one of two reasons. Either because of $I$ does not satisfy the body of $r$ and does not satisfy any of the boolean combinations that appear in the head of $r$. Or because $I$ does not satisfy the body of $r$.

\begin{definition} Let $\Pi = R_{gen} \cup R_{pref}$ be a ground fuzzy answer set optimization program, $I_1, I_2$ be two fuzzy answer sets of $R_{gen}$, $r$ be a fuzzy preference rule in $R_{pref}$, and $C_i$ be boolean combination appearing in $head(r)$. Then, $I_1$ is strictly preferred over $I_2$ w.r.t. $C_i$, denoted by $I_1 \succ_i I_2$, iff $I_1 \models C_i$ and $I_2 \nvDash C_i$ or $I_1 \models C_i$ and $I_2 \models C_i$ and one of the following holds:

\begin{itemize}

\item $C_i = l:\mu$ implies $I_1 \succ_i I_2$ iff $I_1(l) > I_2(l)$.

\item $C_i = not \; l:\mu$ implies $I_1 \succ_i I_2$ iff $I_1(l) < I_2(l)$ or \\ $l$ is undefined in $I_1$ but defined in $I_2$.

\item $C_i = C_{i_1} \wedge C_{i_2}$ implies $I_1 \succ_i I_2$ iff there exists $t \in \{{i_1}, {i_2}\}$ such that $I_1 \succ_t I_2$ and for all other $t' \in \{{i_1}, {i_2}\}$, we have $I_1 \succeq_{t'} I_2$.

\item $C_i = C_{i_1} \vee C_{i_2}$ implies $I_1 \succ_i I_2$ iff there exists $t \in \{{i_1}, {i_2}\}$ such that $I_1 \succ_t I_2$ and for all other $t' \in \{{i_1}, {i_2}\}$, we have $I_1 \succeq_{t'} I_2$.


\end{itemize}
We say, $I_1$ and $I_2$ are equally preferred w.r.t. $C_i$, denoted by $I_1 =_{i} I_2$, iff $I_1 \nvDash C_i$ and $I_2 \nvDash C_i$ or $I_1 \models C_i$ and $I_2 \models C_i$ and one of the following holds:

\begin{itemize}

\item $C_i = l:\mu$ implies $I_1 =_{i} I_2$ iff $I_1(l) = I_2(l)$.

\item $C_i = not \; l:\mu$ implies $I_1 =_{i} I_2$  iff $I_1(l) = I_2(l)$ or \\ $l$ is undefined in both $I_1$ and $I_2$.

\item $C_i = C_{i_1} \wedge C_{i_2}$ implies $I_1 =_{i} I_2$ iff
\[\forall \: t \in \{{i_1}, {i_2}\}, \; I_1 =_{t} I_2.\]

\item $C_i = C_{i_1} \vee C_{i_2}$ implies $I_1 =_{i} I_2$ iff
\[
|\{I_1 \succeq_{t} I_2 \: | \: \forall \: t \in \{{i_1}, {i_2}\} \}| = | \{ I_2 \succeq_{t} I_1 \: | \: \forall \: t \in \{{i_1}, {i_2}\} \}|.
\]

\end{itemize}
We say, $I_1$ is at least as preferred as $I_2$ w.r.t. $C_i$, denoted by $I_1 \succeq_i I_2$, iff $I_1 \succ_i I_2$ or $I_1 =_i I_2$.
\label{def:compination}
\end{definition}

\begin{definition} Let $\Pi = R_{gen} \cup R_{pref}$ be a ground fuzzy answer set optimization program, $I_1, I_2$ be two fuzzy answer sets of $R_{gen}$, $r$ be a fuzzy preference rule in $R_{pref}$, and $C_l$ be boolean combination appearing in $head(r)$. Then, $I_1$ is strictly preferred over $I_2$ w.r.t. $r$, denoted by $I_1 \succ_r I_2$, iff one of the following holds:
\begin{itemize}
\item $I_1 \models_{i} r$ and $I_2 \models_{j} r$ and $i < j$, \\
where $i = \min \{l \; | \; I_1 \models_l r \}$ and $j = \min \{l \; | \; I_2 \models_l r \}$.

\item $I_1 \models_{i} r$ and $I_2 \models_{i} r$ and $I_1 \succ_i I_2$, \\
where $i = \min \{l \; | \; I_1 \models_l r \} = \min \{l \; | \; I_2 \models_l r \}$.

\item $I_1 \models_{i} r$ and $I_2 \models_{irr} r$.
\end{itemize}
We say, $I_1$ and $I_2$ are equally preferred w.r.t. $r$, denoted by $I_1 =_{r} I_2$, iff one of the following holds:
\begin{itemize}
\item $I_1 \models_{i}  r$ and $I_2 \models_{i} r$ and $I_1 =_i I_2$, \\
where $i = \min \{l \; | \; I_1 \models_l r \} = \min \{l \; | \; I_2 \models_l r \}$.
\item $I_1 \models_{irr}  r$ and $I_2 \models_{irr} r$.
\end{itemize}
We say, $I_1$ is at least as preferred as $I_2$ w.r.t. $r$, denoted by $I_1 \succeq_{r} I_2$, iff $I_1 \succ_{r} I_2$ or $I_1 =_{r} I_2$.
\label{def:pref-rule}
\end{definition}
The above definitions specify how fuzzy answer sets are ranked according to a given boolean combination and according to a fuzzy preference rule. Definition \ref{def:compination} shows the ranking of fuzzy answer sets with respect to a boolean combination. However, Definition \ref{def:pref-rule} specifies the ranking of fuzzy answer sets according to a fuzzy preference rule. The following definitions determine the ranking of fuzzy answer sets with respect to a set of fuzzy preference rules.

\begin{definition} [Pareto Preference] Let $\Pi = R_{gen} \cup R_{pref}$ be a fuzzy answer set optimization program and $I_1, I_2$ be fuzzy answer sets of $R_{gen}$. Then, $I_1$ is (Pareto) preferred over $I_2$ w.r.t. $R_{pref}$, denoted by $I_1 \succ_{R_{pref}} I_2$, iff there exists at least one fuzzy preference rule $r \in R_{pref}$ such that $I_1 \succ_{r} I_2$ and for every other rule $r' \in R_{pref}$, $I_1 \succeq_{r'} I_2$. We say, $I_1$ and $I_2$ are equally (Pareto) preferred w.r.t. $R_{pref}$, denoted by $I_1 =_{R_{pref}} I_2$, iff for all $r \in R_{pref}$, $I_1 =_{r} I_2$.
\end{definition}

\begin{definition} [Maximal Preference] Let \\ $\Pi = R_{gen} \cup R_{pref}$ be a fuzzy answer set optimization program and $I_1, I_2$ be fuzzy answer sets of $R_{gen}$. Then, $I_1$ is (Maximal) preferred over $I_2$ w.r.t. $R_{pref}$, denoted by $I_1 \succ_{R_{pref}} I_2$, iff
\[
|\{r \in R_{pref} | I_1 \succeq_{r} I_2\}| > |\{r \in R_{pref} | I_2 \succeq_{r} I_1\}|.
\]
We say, $I_1$ and $I_2$ are equally (Maximal) preferred w.r.t. $R_{pref}$, denoted by $I_1 =_{R_{pref}} I_2$, iff
\[
|\{r \in R_{pref} | I_1 \succeq_{r} I_2\}| = | \{r \in R_{pref} | I_2 \succeq_{r} I_1\}|.
\]
\end{definition}
Observe that the Maximal preference relation is more {\em general} than the Pareto preference relation, since the Maximal preference definition {\em subsumes} the Pareto preference relation.

\section{Course Scheduling with Fuzzy Preferences Problem}

In this section, we show that the course scheduling with fuzzy preferences problem, introduced in \cite{Saad_DFLP}, can be easily and intuitively represented and solved in the fuzzy answer set optimization programs framework as follows.

\begin{example} Quoting the course scheduling with fuzzy preferences problem introduced in \cite{Saad_DFLP}, consider there are $n$ different instructors (denoted by $l_1, \ldots, l_n$) who have to be assigned to $n$ different courses (denoted by $c_1, \ldots, c_n$) in $m$ different rooms (denoted by $r_1, \ldots, r_m$) at $k$ different time slots (denoted by $s_1, \ldots, s_k$) under the following constraints. Exactly one course has to be assigned to each instructor. Different courses cannot be taught in the same room at the same time slot. In addition, every instructor preferences in teaching courses is given as a fuzzy set over courses, every instructor preferences in time slots is given as a fuzzy set over time slots, and every instructor preferences in rooms is given as a fuzzy set over rooms. This course scheduling with fuzzy preferences problem can be represented as a fuzzy answer set optimization program $\Pi = R_{gen} \cup R_{pref}$, where $R_{gen}$ is a set of disjunctive fuzzy logic rules with fuzzy answer set semantics of the form:

\begin{eqnarray}
teaches(l_i, c_1) \hspace{-0.5cm} && : \mu_{i,1} \; \vee \;  teaches(l_i, c_2): \mu_{i,2} \; \vee  \ldots \vee \notag \\ && teaches(l_i, c_n): \mu_{i,n}  \leftarrow
\\
in(r_1, C) \hspace{-0.5cm} && : \nu_{i,1} \; \vee \; in(r_2, C): \nu_{i,2} \; \vee  \ldots \vee \notag \\
&& in(r_m, C): \nu_{i,m}  \leftarrow teaches(l_i, C) : V, \notag\\
&& course(C):1. \label{rule:course-in}
\\
at(s_1, C) \hspace{-0.5cm} && :  v_{i,1} \; \vee \;  at(s_2, C): v_{i,2}  \; \vee \ldots  \vee \notag \\
&& at(s_k, C): v_{i,k} \leftarrow teaches(l_i, C) : V, \notag \\
&& course(C):1. \label{rule:course-at}
\\
inconsistent \hspace{-0.5cm} && :  1   \leftarrow  not \; inconsistent:1, \notag \\
&& teaches(I_1, C):V_1, teaches(I_2, C):V_2, \notag \\
&& I_1 \neq I_2. \label{rule:course-constr1}
\\
inconsistent \hspace{-0.5cm} && :  1  \leftarrow  not \; inconsistent:1, in(R, C):V_1, \notag \\
&& in(R, C'):V_2, at(S, C):V_3, at(S, C'):V_4, \notag \\
&& C \neq C'. \label{rule:course-constr2}
\end{eqnarray}
where $V, V_1, \ldots, V_4$ are annotation variables act as place holders and for all, ($1 \leq i \leq  n$), $teaches(l_i, c_j): \mu_{i,j}$ represents that instructor $l_i$ preference in teaching course $c_j$ is described by the grade membership value $\mu_{i,j}$; $in(r_j, C): \nu_{i,j}$ represents that instructor $l_i$ preference in teaching in room $r_j$ a course $C$ is described by the grade membership value $\nu_{i,j}$; and $at(s_j, C): v_{i,j}$ represents that instructor $l_i$ preference in teaching at time slot $s_j$ a course $C$ is described by the grade membership value $v_{i,j}$. Instructors preference over courses, rooms, and time slots are encoded by the first three disjunctive fuzzy logic rules. The last two disjunctive fuzzy logic rules encode the problem constraints which are every instructor is assigned exactly one course and different courses cannot be taught in the same room at the same time.

The set of fuzzy preference rules, $R_{pref}$, of the fuzzy answer optimization program, $\Pi$, description of the course scheduling with fuzzy preferences problem is given by:
\begin{eqnarray}
teaches(l_i, c_1) \hspace{-0.5cm} && : \mu_{i,1} \; \succ \; teaches(l_i, c_2): \mu_{i,2} \; \succ  \ldots \succ \notag \\ && teaches(l_i, c_n): \mu_{i,n}  \leftarrow
\\
in(r_1, C) \hspace{-0.5cm} && : \nu_{i,1}  \; \succ \; in(r_2, C): \nu_{i,2} \; \succ \ldots \succ \notag \\
&& in(r_m, C): \nu_{i,m} \leftarrow teaches(l_i, C) : V, \notag \notag \\
&& course(C):1. \label{rule:course-pref-in}
\\
at(s_1, C) \hspace{-0.5cm} && :  v_{i,1}   \; \succ \;  at(s_2, C): v_{i,2}  \; \succ \ldots  \succ  \notag \\
&& at(s_k, C): v_{i,k} \leftarrow teaches(l_i, C) : V, \notag \\
&& course(C):1. \label{rule:course-pref-at}
\end{eqnarray}
where for all ($1 \leq i \leq  n$), $\mu_{i,1} \geq \mu_{i,2} \geq \ldots \geq \mu_{i,n}$, similarly, $\nu_{i,1} \geq \nu_{i,2} \geq \ldots \geq \nu_{i,n}$, and $v_{i,1} \geq v_{i,2} \geq \ldots \geq v_{i,n}$.
\label{ex:schedule}
\end{example}

Nevertheless, the fuzzy preference rules, $R_{pref}$, of the fuzzy answer set optimization program, $\Pi$, encoding of the course scheduling with fuzzy preferences problem can be easily and intuitively adapted according to the instructors preferences in many and very flexible ways. For example, as mentioned earlier in the introduction, it can be the case that instructor $i$ is neutral regarding teaching courses $c_1$ and $c_2$ with grade membership value $0.3$ each. This means that $c_1$ and $c_2$ are equally preferred to instructor $i$. Thus, this situation can be represented in instructor $i$ fuzzy preference rule in $R_{pref}$ as
\begin{equation*}
teaches(l_i, c_1): 0.3  \vee teaches(l_i, c_2): 0.3 \leftarrow.
\end{equation*}
Furthermore, although instructor $i$ is neutral regarding teaching courses $c_1$ and $c_2$ with grade membership value $0.3$ each, it can be the case that instructor $i$ has more appeal in teaching course $c_1$ over $c_2$. So that this situation can be intuitively represented in instructor $i$ fuzzy preference rule in $R_{pref}$ as
\begin{equation*}
teaches(l_i, c_1): 0.3 \; \succ \; teaches(l_i, c_2): 0.3 \leftarrow.
\end{equation*}
Moreover, it can be the case that each instructor, $i$, has different rooms preferences and different time slots preferences per each course, $c_j$, as some courses, $c_j$, may require rooms with special equipments installed and/or better to be taught at certain time slots over the other time slots. This can be easily and intuitively achieved by replacing the disjunctive fuzzy logic rules (\ref{rule:course-in}) and (\ref{rule:course-at}) in $R_{gen}$ by the following set of disjunctive fuzzy logic rules for each instructor $l_i$ and for each course $c_j$ as

\begin{eqnarray}
in(r_1, c_j) \hspace{-0.5cm} && : \nu_{i,1} \; \vee \; in(r_2, c_j): \nu_{i,2} \; \vee  \ldots \vee \notag
\\ && in(r_m, c_j): \nu_{i,m}  \leftarrow teaches(l_i, c_j) : V
\\
at(s_1, c_j) \hspace{-0.5cm} && :  v_{i,1} \; \vee \;  at(s_2, c_j): v_{i,2}  \; \vee \ldots  \vee \notag \\
&& at(s_k, c_j): v_{i,k} \leftarrow teaches(l_i, c_j) : V
\end{eqnarray}
In addition to replacing the fuzzy preference rules (\ref{rule:course-pref-in}) and (\ref{rule:course-pref-at}) in $R_{pref}$ by the following set of fuzzy preference rules for each instructor $l_i$ and for each course $c_j$ as
\begin{eqnarray}
in(r_1, c_j) \hspace{-0.5cm} && : \nu_{i,1}  \; \succ \; in(r_2, c_j): \nu_{i,2} \; \succ   \ldots \succ \notag \\
&& in(r_m, c_j): \nu_{i,m} \leftarrow teaches(l_i, c_j) : V
\\
at(s_1, c_j) \hspace{-0.5cm} && :  v_{i,1}   \; \succ \;  at(s_2, c_j): v_{i,2}  \; \succ \ldots  \succ  \notag \\
&& at(s_k, c_j): v_{i,k} \leftarrow teaches(l_i, c_j) : V
\end{eqnarray}
This shows in general that fuzzy answer set optimization programs can be intuitively and flexibly used to represent and reason in the presence of both quantitative and qualitative preferences. This is more clarified by the following instance of the course scheduling with fuzzy preferences problem described below.

\begin{example} Quoting \cite{Saad_DFLP}, assume that two different courses, denoted by $c_1, c_2$, need to be assigned to two different instructors, named $i_1, i_2$, given that only one room, denoted by $r_1$, is available and two different time slots, denoted by $s_1, s_2$ are allowed, with fuzzy preferences as described below. This instance of the course scheduling with fuzzy preferences problem can be encoded as an instance of the fuzzy answer set optimization program, $\Pi = R_{gen} \cup R_{pref}$, presented in Example \ref{ex:schedule}, as a fuzzy answer set optimization program, $\Pi' = R'_{gen} \cup R'_{pref}$, where in addition to the last two disjunctive fuzzy logic rules, (\ref{rule:course-constr1}) and (\ref{rule:course-constr2}), of $R_{gen}$ in $\Pi$ described in Example \ref{ex:schedule}, $R'_{gen}$ also contains the following disjunctive fuzzy logic rules:

\begin{eqnarray*}
teaches(i_1, c_1) \hspace{-0.5cm} && : 0.9 \; \vee \; teaches(i_1, c_2): 0.5 \leftarrow  \\ 
teaches(i_2, c_1) \hspace{-0.5cm} && : 0.4 \; \vee \; teaches(i_2, c_2): 0.7 \leftarrow  \\ 
in(r_1, C) \hspace{-0.5cm} && : 0.8  \leftarrow teaches(i_1, C) : V, course(C):1.  \\
in(r_1, C) \hspace{-0.5cm} && : 0.3  \leftarrow teaches(i_2, C) : V, course(C):1. \\
at(s_1, C) \hspace{-0.5cm} && : 0.5 \; \vee \;  at(s_2, C): 0.5 \leftarrow  teaches(i_1, C) : V, \\
&& course(C):1.  \\
at(s_1, C) \hspace{-0.5cm} && : 0.9 \; \vee \;  at(s_2, C): 0.2 \leftarrow  teaches(i_2, C) : V, \\
&& course(C):1.  \\
course(c_1) \hspace{-0.5cm} && :1 \leftarrow \\
course(c_2) \hspace{-0.5cm} && :1 \leftarrow
\end{eqnarray*}
In addition, $R'_{pref}$, contains the fuzzy preference rules:
\begin{eqnarray*}
&& teaches(i_1, c_1): 0.9 \; \succ \; teaches(i_1, c_2): 0.5 \leftarrow  \\
&& teaches(i_2, c_2): 0.7 \; \succ \;  teaches(i_2, c_1): 0.4 \leftarrow  \\
&& at(s_1, C): 0.5 \; \vee \;  at(s_2, C): 0.5 \leftarrow  teaches(i_1, C) : V  \\
&& at(s_1, C): 0.9 \; \succ \;  at(s_2, C): 0.2 \leftarrow  teaches(i_2, C) : V  \\
&& in(r_1, C): V  \leftarrow teaches(I, C) : V'
\end{eqnarray*}
The ground instantiation of the fuzzy preference rules in  $R'_{pref}$ have ten relevant ground fuzzy preference rules which are:
\begin{eqnarray*}
r_1: teaches(i_1, c_1): 0.9 \; \succ \; teaches(i_1, c_2): 0.5 \leftarrow  \\
r_2:  teaches(i_2, c_2): 0.7 \; \succ \;  teaches(i_2, c_1): 0.4 \leftarrow  \\
r_3: at(s_1, c_1): 0.5  \vee  at(s_2, c_1): 0.5 \leftarrow  teaches(i_1, c_1) : 0.9  \\
r_4: at(s_1, c_2): 0.5  \vee  at(s_2, c_2): 0.5 \leftarrow  teaches(i_1, c_2) : 0.5  \\
r_5: at(s_1, c_1): 0.9  \succ  at(s_2, c_1): 0.2 \leftarrow  teaches(i_2, c_1) : 0.4  \\
r_6: at(s_1, c_2): 0.9  \succ  at(s_2, c_2): 0.2 \leftarrow  teaches(i_2, c_2) : 0.7  \\
r_7: in(r_1, c_1): 0.8  \leftarrow teaches(i_1, c_1) : 0.9  \\
r_8: in(r_1, c_2): 0.8  \leftarrow teaches(i_1, c_2) : 0.5  \\
r_9: in(r_1, c_1): 0.3  \leftarrow teaches(i_2, c_1) : 0.4  \\
r_{10}: in(r_1, c_2): 0.3  \leftarrow teaches(i_2, c_2) : 0.7
\end{eqnarray*}
\end{example}
The generator rules, $R'_{gen}$, of the fuzzy answer set optimization program, $\Pi'$, has four fuzzy answer sets that are:
\[
\begin{array}{l}
I_1 = \{ \; teaches(i_1,c_1): 0.9,\; teaches(i_2,c_2):0.7, \\
\qquad \;\;\; at(s_1,c_1):0.5, \; at(s_2,c_2):0.2, \; in(r_1,c_1):0.8, \\
\qquad \;\;\; in(r_1,c_2):0.3, \; course(c_1):1, \; course(c_2):1 \; \}
\end{array}
\]
\[
\begin{array}{l}
I_2 = \{ \; teaches(i_1,c_1):0.9, \; teaches(i_2,c_2):0.7, \\
\qquad \;\;\; at(s_2,c_1):0.5, \; at(s_1,c_2):0.9, \; in(r_1,c_1):0.8, \\
\qquad \;\;\; in(r_1,c_2):0.3, \; course(c_1):1, \; course(c_2): 1 \; \}
\end{array}
\]
\[
\begin{array}{l}
I_3 = \{ \; teaches(i_1,c_2):0.5, \; teaches(i_2,c_1):0.4,  \\
\qquad \;\;\; at(s_2,c_2):0.5, \; at(s_1,c_1):0.9, \; in(r_1,c_2):0.8, \\
\qquad \;\;\; in(r_1,c_1):0.3, \; course(c_1):1, \; course(c_2): 1 \; \}
\end{array}
\]
\[
\begin{array}{l}
I_4 = \{ \; teaches(i_1,c_2):0.5, \; teaches(i_2,c_1):0.4, \\
\qquad \;\;\; at(s_1,c_2):0.5, \; at(s_2,c_1):0.2, \; in(r_1,c_2):0.8, \\
\qquad \;\;\; in(r_1,c_1):0.3, \; course(c_1):1, \; course(c_2): 1 \; \}
\end{array}
\]
%
%
We can easily verify that
\[
\begin{array}{llll}
I_1 \models_{1} r_1, & \; I_1 \models_{1} r_2, & \;  I_1 \models_{1} r_3, & \;  I_1 \models_{irr} r_4, \\
I_1 \models_{irr} r_5, & \; I_1 \models_{2} r_6, & \;  I_1 \models_{1} r_7, & \;  I_1 \models_{irr} r_8, \\
& \; I_1 \models_{irr} r_9, & \;  I_1 \models_{1} r_{10}.
\\
\\
I_2 \models_{1} r_1, & \; I_2 \models_{1} r_2, & \;  I_2 \models_{1} r_3, & \;  I_2 \models_{irr} r_4, \\
I_2 \models_{irr} r_5, & \; I_2 \models_{1} r_6, & \;  I_2 \models_{1} r_7, & \;  I_2 \models_{irr} r_8, \\
& \; I_2 \models_{irr} r_9, & \;  I_2 \models_{1} r_{10}.
\\
\\
I_3 \models_{2} r_1, & \; I_3 \models_{2} r_2, & \;  I_3 \models_{irr} r_3, & \;  I_3 \models_{1} r_4, \\
I_3 \models_{1} r_5, & \; I_3 \models_{irr} r_6, & \;  I_3 \models_{irr} r_7, & \;  I_3 \models_{1} r_8, \\
& \; I_3 \models_{1} r_9, & \;  I_3 \models_{irr} r_{10}.
\\
\\
I_4 \models_{2} r_1, & \; I_4 \models_{2} r_2, & \;  I_4 \models_{irr} r_3, & \;  I_4 \models_{1} r_4, \\
I_4 \models_{2} r_5, & \; I_4 \models_{irr} r_6, & \;  I_4 \models_{irr} r_7, & \;  I_4 \models_{1} r_8, \\
& \; I_4 \models_{1} r_9, & \;  I_4 \models_{irr} r_{10}.
\end{array}
\]
Therefore, $I_2$ is the top (Maximal) preferred fuzzy answer set and $I_4$ is the least (Maximal) preferred fuzzy answer set. However, $I_1$ is (Maximal) preferred over $I_3$ and $I_3$ is (Maximal) preferred over $I_4$ as well as $I_2$ is (Maximal) preferred over $I_1$. Thus, the ranking of the fuzzy answer sets from the top (Maximal) preferred fuzzy answer set to the least (Maximal) preferred fuzzy answer set is $I_2, I_ 1, I_3, I_4$, i.e., \\ $I_2 \quad \succ_{R_{pref}} I_ 1 \quad \succ_{R_{pref}} I_3 \quad \succ_{R_{pref}} I_4$.

\section{Implementation}

In this section, we provide an implementation for fuzzy answer set optimization programs in fuzzy answer programming. This is achieved by providing a translation from a fuzzy answer set optimization program, $\Pi = R_{gen} \cup R_{pref}$, into a fuzzy answer set program, $\Pi^e = R^e_{gen} \cup R^e_{pref}$ \cite{Saad_DFLP,Saad_eflp,Subrahmanian_B}, where $R^e_{gen} = R_{gen}$ and $R_{pref}$ is translated into a set of extended fuzzy logic rules \cite{Saad_eflp}, where the fuzzy answer sets of $\Pi$ are equivalent to the fuzzy answer sets of $\Pi^e$. The syntax and semantics of a set of extended fuzzy logic rules \cite{Saad_eflp} is the same as the syntax and semantics of a set of disjunctive fuzzy logic rules presented earlier in this paper except that; extended fuzzy logic rules allow one fuzzy annotated literal in the head of rules and fuzzy answer sets of a set of extended fuzzy logic rules can be partial mappings.

Let without loss of generality, $r$, be a fuzzy preference rule of the form
\begin{eqnarray*}
C_1 \succ C_2 \succ \ldots \succ C_k \leftarrow l_{k+1}:\mu_{k+1},\ldots, l_m:\mu_m, \notag \\
not\; l_{m+1}:\mu_{m+1},\ldots, not\;l_n:\mu_n.
\end{eqnarray*}
where each $C_i$ in the head of $r$ is represented as a generalized fuzzy annotated DNF, that is, of the form
\[
(s_{1,1}:\mu_{1,1} \land \ldots s_{1,t_1}:\mu_{1,t_1} ) \vee \ldots \vee  (s_{u,1}:\mu_{u,1} \land \ldots s_{u,t_u}:\mu_{u,t_u})
\]
where each $s_{v,w}:\mu_{v,w}$ is a fuzzy annotated literal possibly proceeded by non-monotonic negation. In addition, let $sat(r, i)$ be a predicate denoting that the boolean combination $C_i$ in the head of a fuzzy preference rule $r$ is satisfied.

The translation of the fuzzy answer set optimization program, $\Pi = R_{gen} \cup R_{pref}$, into a fuzzy answer set program, $\Pi^e = R^e_{gen} \cup R^e_{pref}$, proceeds as follows, where $R^e_{gen} = R_{gen}$ and $R^e_{pref}$ contains the following extended fuzzy logic rules:

\begin{itemize}

\item For each fuzzy preference rule, $r \in R_{pref}$, we have in $R^e_{pref}$ the extended fuzzy logic rule
\begin{eqnarray*}
body(r):1 \leftarrow l_{k+1}:\mu_{k+1},\ldots, l_m:\mu_{m},  \notag \\
not\; l_{m+1}:\mu_{m+1}, \ldots, not\;l_{n}:\mu_{n} \label{rule:body}
\end{eqnarray*}

\item For each boolean combination in the head of a fuzzy preference rule, $r$, of the form
\begin{eqnarray*}
C_i = (s_{1,1}:\mu_{1,1} \land \ldots s_{1,t_1}:\mu_{1,t_1} ) \vee \ldots \vee  \\
(s_{u,1}:\mu_{u,1} \land \ldots s_{u,t_u}:\mu_{u,t_u})
\end{eqnarray*}
we have in $R^e_{pref}$ the extended fuzzy logic rules:

\begin{eqnarray*}
sat(r, i):1 & \leftarrow & s_{1,1}:\mu_{1,1}, \ldots , s_{1,t_1}:\mu_{1,t_1}, \\
&& body(r):1 \\
& \ldots \ldots & \\
sat(r, i):1 & \leftarrow & s_{u,1}:\mu_{u,1}, \ldots, s_{u,t_u}:\mu_{u,t_u}, \\
&& body(r):1
\end{eqnarray*}

\item For each fuzzy preference rule, $r \in R_{pref}$, we have in $R^e_{pref}$ the extended fuzzy logic rules

\begin{eqnarray*}
sat(r, irr):1 & \leftarrow & not \; body(r):1 \label{rule:irr-body} \\
sat(r, irr):1 & \leftarrow & not \; sat(r, 1):1, \ldots, not \; sat(r, k):1, \\
&& body(r):1 \label{rule:irr}
\end{eqnarray*}
\end{itemize}
Obviously, the fuzzy answer sets of $\Pi$ are in one-to-one correspondence to the fuzzy answer sets of $\Pi^e$.

\begin{theorem} Let $\Pi = R_{gen} \cup R_{pref}$ be a fuzzy answer set optimization program, $\Pi^e = R^e_{gen} \cup R^e_{pref}$ be the fuzzy answer program translation of $\Pi$, $r$ be a fuzzy preference rule in $R_{pref}$, $C_i$ be a boolean combination in the head of $r$, $I$ be a fuzzy answer set of $\Pi$, and $I'$ be a fuzzy answer set of $\Pi^e$ that corresponds to $I$. Then,
\begin{itemize}
\item $I \models_{i} r$ iff $I' \models sat(r, i):1$.

\item $I \models_{irr} r$ iff $I' \models sat(r, irr):1$.
\end{itemize}
\label{thm:main}
\end{theorem}

%
%
%

Moreover, we show that the fuzzy answer set optimization programs syntax and semantics naturally subsume and generalize the classical answer set optimization programs syntax and semantics \cite{ASO} under the Pareto preference relation, since there is no notion of Maximal preference relation has been defined for the classical answer set optimization programs.

A classical answer set optimization program, $\Pi^c$, consists of two separate classical logic programs; a classical answer set program, $R^c_{gen}$, and a classical preference program, $R^c_{pref}$ \cite{ASO}. The first classical logic program, $R^c_{gen}$, is used to generate the classical answer sets. The second classical logic program, $R^c_{pref}$, defines classical context-dependant preferences that are used to form a preference ordering among the classical answer sets of $R^c_{gen}$.

Any classical answer set optimization program, $\Pi^c = R^c_{gen} \cup R^c_{pref}$, can be represented as a fuzzy answer set optimization program, $\Pi = R_{gen} \cup R_{pref}$, where all fuzzy annotations appearing in every fuzzy logic rule in $R_{gen}$ and all fuzzy annotations appearing in every fuzzy preference rule in $R_{pref}$ are equal to $1$, which means the truth value {\em true}. For example, for a classical answer set optimization program, $\Pi^c = R^c_{gen} \cup R^c_{pref}$, that is represented by the fuzzy answer set optimization program, $\Pi = R_{gen} \cup R_{pref}$, the classical logic rule
\begin{eqnarray*}
a_1 \; \vee \ldots \vee \; a_k \leftarrow  a_{k+1}, \ldots, a_m, not\; a_{m+1},
\ldots, not\;a_{n}
\end{eqnarray*}
is in $R^c_{gen}$, where $\forall (1 \leq i \leq n)$, $a_i$ is an atom, iff
\begin{eqnarray*}
a_1:1 \; \vee \ldots \vee \; a_k:1 \leftarrow  a_{k+1}:1, \ldots, a_m:1, \\ not\; a_{m+1}:1,
\ldots, not\;a_{n}:1
\end{eqnarray*}
is in $R_{gen}$. It is worth noting that the syntax and semantics of this class of fuzzy answer set programs is the same as the syntax and semantics of the classical answer set programs \cite{Saad_DFLP,Saad_eflp}. In addition, the classical preference rule
\begin{eqnarray*}
C_1 \succ C_2 \succ \ldots \succ C_k \leftarrow l_{k+1},\ldots, l_m,
not\; l_{m+1},\ldots, not\;l_{n}
\end{eqnarray*}
is in $R^c_{pref}$, where $l_{k+1}, \ldots, l_{n}$ are literals and $C_1, C_2, \ldots, C_k$ are boolean combinations over a set of literals, iff
\begin{eqnarray*}
C_1 \succ C_2 \succ \ldots \succ C_k \leftarrow l_{k+1}:1,\ldots, l_m:1, \\
not\; l_{m+1}:1,\ldots, not\;l_{n}:1
\end{eqnarray*}
is in $R_{pref}$ and every literal appearing in $C_1, C_2, \ldots, C_k$ is annotated with the fuzzy annotation constant $1$.

The following results show that the syntax and semantics of the fuzzy answer set optimization programs subsume the syntax and semantics of the classical answer set optimization programs \cite{ASO}, assuming that \cite{ASO} assigns the lowest rank to the classical answer sets that do not satisfy either the body of a classical preference rule or the body of a classical preference and any of the boolean combinations appearing in the head of the classical preference rule.

\begin{theorem} Let $\Pi = R_{gen} \cup R_{pref}$ be the fuzzy answer set optimization program equivalent to a classical answer set optimization program, $\Pi^c = R^c_{gen} \cup R^c_{pref}$. Then, the preference ordering of the fuzzy answer sets of $R_{gen}$ w.r.t. $R_{pref}$ coincides with the preference ordering of the classical answer sets of $R^c_{gen}$ w.r.t. $R^c_{pref}$.
\label{thm:1}
\end{theorem}

\begin{theorem} Let $\Pi = R_{gen} \cup R_{pref}$ be the fuzzy answer set optimization program equivalent to a classical answer set optimization program, $\Pi^c = R^c_{gen} \cup R^c_{pref}$. A fuzzy answer set $I$ of $R_{gen}$ is Pareto preferred fuzzy answer set w.r.t. $R_{pref}$ iff a classical answer set $I^c$ of $R^c_{gen}$, equivalent to $I$, is Pareto preferred classical answer set w.r.t. $R^c_{pref}$.
\label{thm:2}
\end{theorem}
Theorem \ref{thm:1} shows in general that fuzzy answer set optimization programs can be used only for representing and reasoning about qualitative preferences under the classical answer set programming framework, under both Maximal and Pareto preference relations, by simply replacing any fuzzy annotation appearing in a fuzzy answer set optimization program by the constant fuzzy annotation $1$. However, Theorem \ref{thm:2} shows the subsumption result of the classical answer set optimization programs.

\section{Conclusions and Related Work}

We developed syntax and semantics of a logical framework for representing and reasoning about both quantitative and qualitative preferences in a unified logical framework, namely fuzzy answer set optimization programs. The proposed framework is necessary to allow representing and reasoning in the presence of both quantitative and qualitative preferences across fuzzy answer sets. This is to allow the ranking of the fuzzy answer sets from the top preferred fuzzy answer set to the least preferred fuzzy answer set, where the top preferred fuzzy answer set is the one that is most desirable. Fuzzy answer set optimization programs modify and generalize the classical answer set optimization programs described in \cite{ASO}. We have shown the application of fuzzy answer set optimization programs to the course scheduling with fuzzy preferences problem described in \cite{Saad_DFLP}. In addition, we provided an implementation for fuzzy answer set optimization programs in fuzzy answer set programming.

To the best of our knowledge, this development is the first to consider a logical framework for reasoning about quantitative preferences, in general, and reasoning about both quantitative and qualitative preferences in particular. However, qualitative preferences were introduced in classical answer set programming in various forms. In \cite{Schaub-Comp}, qualitative preferences are defined among the rules of classical logic programs, whereas qualitative preferences among the literals described by the classical logic programs are introduced in \cite{Sakama}. Classical answer set optimization \cite{ASO} and classical logic programs with ordered disjunctions \cite{LPOD} are two classical answer set programming based qualitative preference handling approaches, where context-dependant qualitative preferences are defined among the literals specified by the classical logic programs. Application-dependant qualitative preference handling approaches for planning were presented in \cite{Son-Pref,Schaub-Pref07}, where qualitative preferences among actions, states, and trajectories are defined, which are based on temporal logic. The major difference between \cite{Son-Pref,Schaub-Pref07} and \cite{ASO,LPOD} is that the former are specifically developed for planning, but the latter are application-independent.

Contrary to the existing approaches for reasoning about qualitative preferences in classical answer set programming, where qualitative preference relations are specified among rules and literals in one classical logic program, a classical answer set optimization program consists of two separate classical logic programs; a classical answer set program and a qualitative preference program \cite{ASO}. The classical answer set program is used to generate the classical answer sets and the qualitative preference program defines context-dependant qualitative preferences that are used to form a qualitative preference ordering among the classical answer sets generated by the classical answer set program.

Following \cite{ASO}, fuzzy answer set optimization programs presented in this paper distinguish between fuzzy answer sets generation and fuzzy preference based fuzzy answer sets evaluation, which has several advantages. In particular, the set of fuzzy preference rules, in a fuzzy answer set optimization program, is specified independently from the type of fuzzy logic rules used to generate the fuzzy answer sets in the fuzzy answer set optimization program, which makes preference elicitation easier and the whole approach more intuitive and easy to use in practice. In addition, more expressive forms of fuzzy preferences can be represented in fuzzy answer set optimization programs, since they allow several forms of boolean combinations in the heads of the fuzzy preference rules.

In \cite{Saad_ASOG}, the classical answer set optimization programs have been extended to allow classical aggregate preferences. The introduction of classical aggregate preferences to classical answer set optimization programs have made the encoding of multi-objectives optimization problems and Nash equilibrium strategic games more intuitive and easy. The syntax and semantics of the classical answer set optimization programs with classical aggregate preferences were based on the syntax and semantics of classical answer set optimization programs \cite{ASO} and classical aggregates in classical answer set programming \cite{Recur-aggr}. It has been shown in \cite{Saad_ASOG} that the syntax and semantics of classical answer set optimization programs with classical aggregate preferences subsumes the syntax and semantics of classical answer set optimization programs described in \cite{ASO}.

\bibliographystyle{named}
\bibliography{Saad13LFP}

\end{document}